\documentclass[10pt,twocolumn,letterpaper]{article}

\usepackage[pagenumbers]{cvpr} %

\usepackage[accsupp]{axessibility} %
\usepackage{tabularx}
\usepackage{amsmath}
\usepackage{amssymb}
\usepackage{paralist} %
\usepackage{bbm}      %
\usepackage{multirow} %
\usepackage{comment}  %

\newlength{\imageheight}
\newcommand{\tophalf}[1]{
    \settoheight{\imageheight}{\includegraphics{#1}}
    \includegraphics[width=\linewidth, clip, trim=0 0.5\imageheight{} 0 0]{#1}
}
\usepackage[dvipsnames]{xcolor}

\newcommand{\E}{\mathbb{E}}

\renewcommand{\paragraph}[1]{\vspace{2pt plus 1pt minus 1pt}\noindent{\bf #1}\;}

\usepackage{enumitem}
\setlist{itemsep=0.4\parsep,topsep=0.4\parsep, parsep=0pt,partopsep=0pt,leftmargin=1em,wide=0pt}

\definecolor{cvprblue}{rgb}{0.21,0.49,0.74}
\usepackage[pagebackref,breaklinks,colorlinks,citecolor=cvprblue]{hyperref}

\def\paperID{6434} %
\def\confName{CVPR}
\def\confYear{2024}

\title{Rapid Motor Adaptation for Robotic Manipulator Arms}
\author{
Yichao Liang$^{1,2}$
\and
Kevin Ellis$^3$
\and
Jo\~{a}o Henriques$^2$\\
\and
$^1$Computational and Biological Learning Lab, University of Cambridge\\
$^2$Visual Geometry Group, University of Oxford\\
$^3$Cornell University\\
{\tt\small yliang6@gmail.com},
{\tt\small kellis@cornell.edu},
{\tt\small joao@robots.ox.ac.uk}
}

\begin{document}
\maketitle
\begin{abstract}
Developing generalizable manipulation skills is a core challenge in embodied AI. This includes generalization across diverse task configurations, encompassing variations in object shape, density, friction coefficient, and external disturbances such as forces applied to the robot. 
Rapid Motor Adaptation (RMA) offers a promising solution to this challenge.
It posits that essential hidden variables influencing an agent's task performance, such as object mass and shape, can be effectively inferred from the agent's action and proprioceptive history. 
Drawing inspiration from RMA in locomotion and in-hand rotation, we use depth perception to develop agents tailored for rapid motor adaptation in a variety of manipulation tasks.
We evaluated our agents on four challenging tasks from the Maniskill2 benchmark,
namely pick-and-place operations with hundreds of objects from the YCB and EGAD datasets, peg insertion with precise position and orientation, and operating a variety of faucets and handles, with customized environment variations.
Empirical results demonstrate that our agents surpass state-of-the-art methods like automatic domain randomization and vision-based policies, obtaining better generalization performance and sample efficiency.
\end{abstract}    
\section{Introduction}
\label{sec:intro}

\begin{figure*}
  \centering
  \begin{minipage}[b]{\linewidth}
    \centering
    \includegraphics[width=\linewidth]{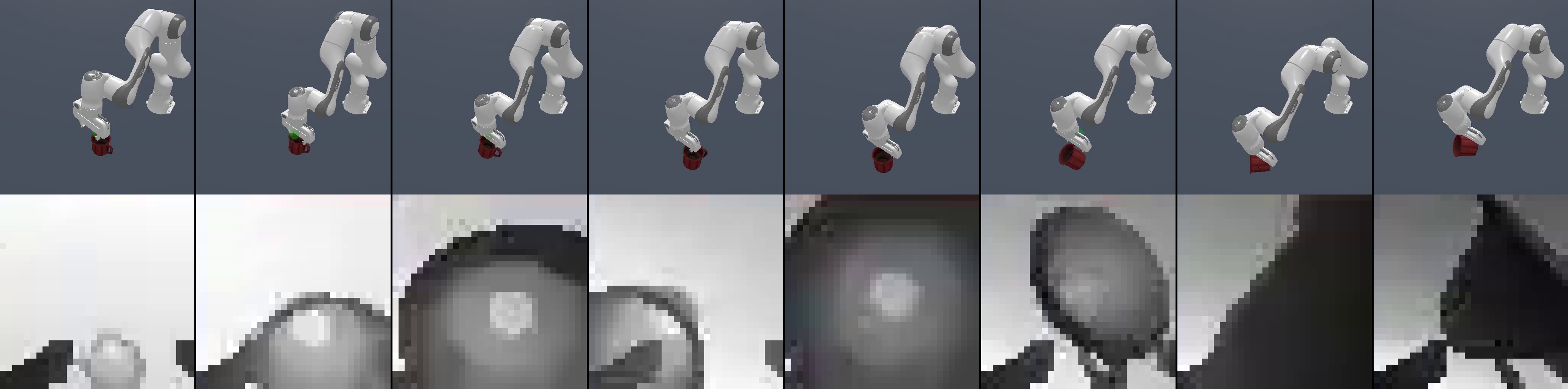}
  \end{minipage}
  \begin{minipage}[b]{\linewidth}
    \centering
    \includegraphics[width=\linewidth]{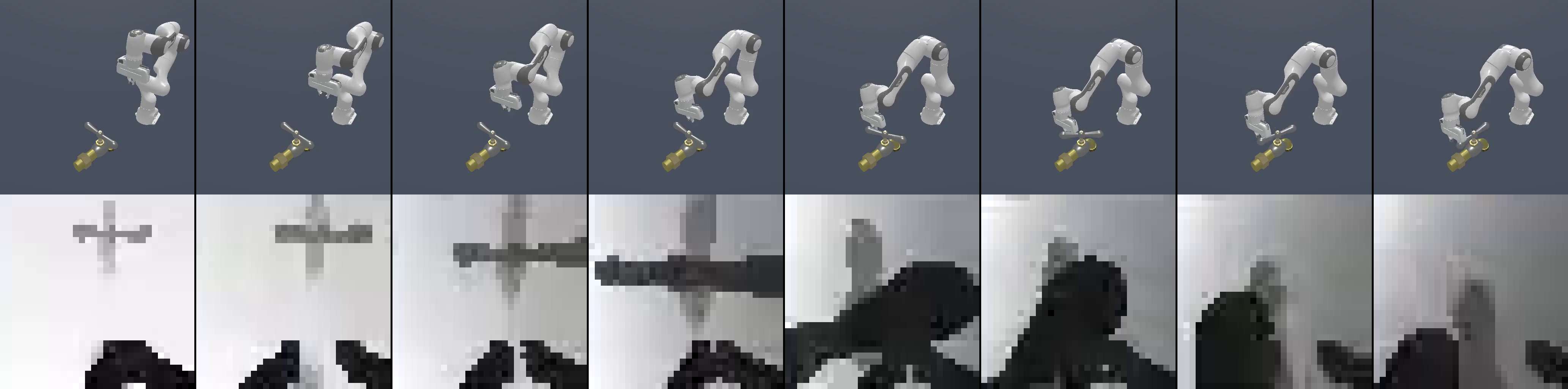}
  \end{minipage}
  \vspace{-12pt}
  \begin{minipage}[b]{\linewidth}
    \centering
    \tophalf{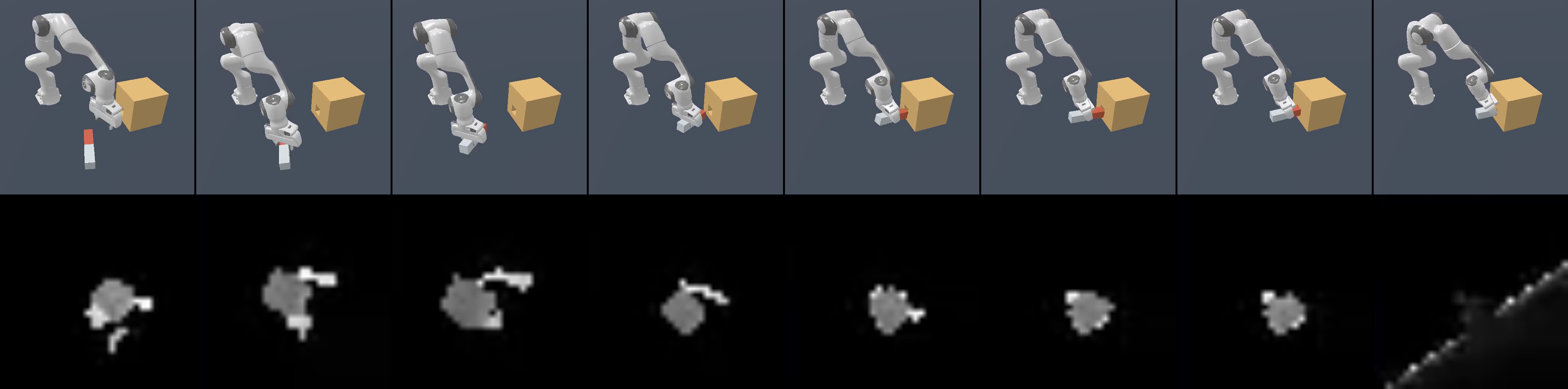}
  \end{minipage}
  \vspace{-12pt}
  \begin{minipage}[b]{\linewidth}
    \centering
    \tophalf{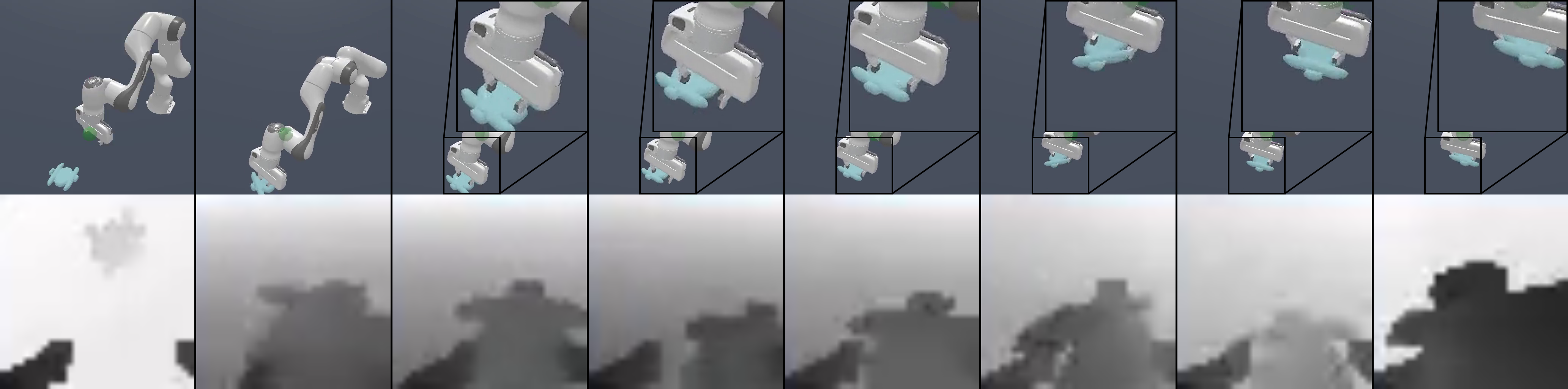}
  \end{minipage}
  \caption{Visualization of an action trajectory by \textit{RMA}$^2$ in each of the four tasks. 
  The top two trajectories also depict the corresponding low-resolution depth images as seen by the adapter module.
  We highlight a few interesting behaviors.
  In the first trajectory, for the Pick \& Place task (YCB dataset), the agent first attempts to pick up a cup by the rim.
  This fails because the rim, in this instance of randomization, is too wide for its gripper.
  The agent then reattempted by grasping it by the handle, which succeeded.
  In the second trajectory, from the Faucet Turning task, we see the agent did not grasp the handle, but only pushed it with one finger to rotate it. The depth image shows the precise positioning of the end effector.
  In the third trajectory, we see the agent did not aim correctly for insertion on the first attempt.
  This is due to the external disturbances applied to the peg, and the fact that the hole has a very small clearance at the level of millimeters.
  But it succeeded after ``jiggling'' the peg around the correct position, a strategy that mimics human behavior.
  In the fourth trajectory, Pick \& Place (EGAD dataset), the agent attempts to pick up a previously-unseen EGAD object.
  The object is too wide for the agent to grasp it from the top, as it lays flat on the floor (a zoomed in inset picture is shown).
  The agent picks up the object by pressing the left side of the object with its left finger and inserting its right finger beneath the object, which is a fair strategy to pick up a flat object.
  }
  \label{fig:trajectories}
\end{figure*}

With recent advances in computer vision \cite{kirillov2023segment, oquab2023dinov2, gothoskar20213dp3} and high-level planning~\cite{ahn2022saycan, wang2023voyager}, dexterous manipulation of objects (i.e. low-level control skills) remains one of the last major obstacles to the creation of robots that can help in general manipulation tasks.
Such an advance would have a wide-ranging impact, allowing robots to take on repetitive tasks in industry and in households.

Classical approaches to robotic manipulation often rely on accurate models of both the robot and the environment~\cite{corke2011robotics}. 
The complexity of creating these models can be a significant hurdle, as they need to account for various physical properties and constraints. 
On the other hand, many Reinforcement Learning (RL) methods are very sample-inefficient, and fail to generalize robustly~\cite{sutton2018reinforcement}.
Many efforts have therefore been invested into simulation training for real-world deployment~\cite{zhao2020sim}.
However, models trained in simulation often fail to perform well in the real world due to the sim-to-real gap -- direct deployment (without any domain adaptation) results in decreased performance~\cite{anderson2021sim}.
More generally, RL agents face the challenge of generalizing to unseen tasks or even tasks with out-of-distribution configurations.

To address these limitations, \citet{kumar2021rma} proposed Rapid Motor Adaptation (RMA), and demonstrated it for quadruped robot locomotion.
The main idea behind RMA is to train a policy that is conditional on environmental factors which are not available in real-world deployment, but are easily randomized and conditioned on during simulation training. A predictor, called the \emph{adaptation module} (adapter for brevity), is then trained to regress these factors from available sensors (such as proprioception).
This is possible because environmental factors, such as the density, friction, and ground elevation, can be reasonably inferred based on the dynamic response of the robot (e.g. the difference between desired and actually observed motion).
In particular, these factors do not usually have to be precisely predicted for the agent to successfully conduct these tasks, which is why a low-dimensional projection of the environment factors is sufficient (and removes the ambiguity of useless but difficult-to-estimate factors)~\citep{kumar2021rma}.
 
While this demonstrates an encouraging path forward, it is not straightforward to bring RMA to general manipulation tasks, which feature diverse objectives and behaviours depending on each object's characteristics. Proprioception alone does not suffice, as it only contains information about the object after touching it -- visual reasoning prior to grasping is required.
We aim to bring the generalization ability of RMA to a broad spectrum of manipulation tasks involving rigid bodies, such as pick-and-place operations, peg insertion, and faucet or lever turning.

We achieve this through several contributions:
\begin{enumerate}
    \item We propose category and instance dictionaries as a strong proxy for geometry-aware manipulation (\cref{sec:cat-inst-dictionaries}), which is crucial to learn policies that are not transferable across objects, e.g. grasping handles in different positions.
    \item We also propose to use a depth convolutional neural network to estimate part of the privileged information about the environment, which performs object category and instance classification only \emph{implicitly} (\cref{sec:adapter}).
    \item As far as we are aware, leveraging these modifications, we are the first to apply rapid motor adaptation to \emph{general} object manipulation tasks with robot arms.
    \item As a smaller contribution, we present a unified formalization of the objectives of the two learning phases of rapid motor adaptation (\cref{eq:policy-phase} and \cref{eq:adapter-phase}), which we believe can be useful in future developments based on this framework.
    \item 
    Through extensive experiments in four Maniskill2 tasks, we demonstrate that our method outperforms several strong baselines, including state-of-the-art techniques with automatic domain randomization \cite{handa2023dextreme, akkaya2019solving} and vision-based policies trained with domain randomization (\cref{sec:result}).
\end{enumerate}

\section{Related Work}
\label{sec:related_work}
Classical control methods have long been the foundation for manipulation tasks \cite{manipulation}.
These approaches, however, usually demand exacting models of both the robots and their operating environments, where even minor discrepancies can lead to performance degradation or task failure. 
Moreover, they face limitations in adapting to object variations in size, weight, and texture, requiring manual recalibration -- a notable hindrance to scalability and flexibility in dynamic real-world applications.

In response, reinforcement learning (RL) with massive compute has emerged as a powerful alternative for learning manipulation skills \citep{fu2016one, gualtieri2018pick, gualtieri2018learning, popov2017data, mahler2017learning, sehgal2019deep, chen2019towards, xiao2019online, li2020acder, pore2020simple, kalashnikov2018scalable}.
Nonetheless, sample efficient generalization remains challenging.
Techniques such as Domain Randomization and Dynamic Randomization \cite{tobin2017domain, sadeghi2016cad2rl, peng2018sim, andrychowicz2020learning} have been adopted widely to leverage massive computational resources to train policies across varied environmental parameters, aiming to cultivate robustness to environmental shifts in a model-agnostic manner.
Subsequent developments have refined this approach, introducing learning and adaptation mechanisms for randomization to enhance sample efficiency and generalization.
For example, \citet{zakharov2019deceptionnet} uses a set of encoder-decoder ``deception'' modules to apply randomization to make the tasks difficult for the policy.
Active Domain Randomization searches for the most informative environment variations within the given randomization ranges, where the informativeness is measured as the discrepancies of policy rollouts in randomized and non-randomized environment instances \citep{mehta2020active}.
Automatic Domain Randomization (ADR) adapts the ranges for the randomization distribution based on the policy performance under the current randomization setting to help improve sample efficiency \cite{handa2023dextreme, akkaya2019solving}.
We take ADR as one of the baselines for comparison.

Rather than having the policy be independent of the environment parameters, we can condition the policy on privileged parameters in simulation, conceptually related to system identification in control theory \cite{karayiannidis2016adaptive, yu2017preparing, yu2018policy}.
For instance, during deployment, physics parameters can be inferred through a trained module \cite{yu2017preparing} or optimized directly by evolutionary algorithms \cite{yu2018policy}.
However, inferring the exact parameters may not always be feasible or optimal for generalization.

Recently, Rapid Motor Adaptation (RMA) has presented a novel approach by learning to predict low-dimensional embeddings of environment parameters, demonstrating remarkably sample-efficient generalization in locomotion and in-hand manipulation tasks \cite{kumar2021rma, qi2023hand}.
In locomotion, an agent trained entirely in simulation was able to traverse through changing terrains, with changing payloads, and with wear and tear, while using solely proprioception.
Building on this, \citet{qi2023hand} extend RMA to robotic in-hand rotation.
They demonstrated that the controller, trained entirely in simulation on only cylindrical objects, can be directly deployed to a real robot hand to rotate dozens of objects with diverse sizes, shapes, and weights over one axis.
Despite its potential, RMA's application to general manipulation, where object states and goals vary from episode to episode, remains non-trivial.

\section{Method}\label{sec:method}

Our work extends RMA \cite{kumar2021rma, qi2023hand} to perform object manipulation with robot arms.
The key novelties are to condition the policy on diverse manipulation goals, and to visually infer object properties from depth images, which requires several modifications.

The main idea of RMA is to train a policy with a high amount of \emph{domain randomization}, which is possible as long as the policy is conditioned on privileged information about the random environment parameters. Then in a second phase, an adapter is learned that estimates the privileged information from readily-available inputs, such as the history of a robot's joints (proprioception). This allows the policy to exhibit highly-specific behaviours for different environments and situations, such as how to deal with low friction or high masses, without direct access to these factors during deployment.
In our work, these factors are extended to include factors relevant to object-manipulation (e.g., what are we manipulating?), and RMA is extended beyond just proprioception, which cannot estimate object-manipulation factors prior to robot-object contact.

\subsection{Policy Training Phase}

\begin{figure}
    \centering
    \includegraphics[width=\columnwidth]{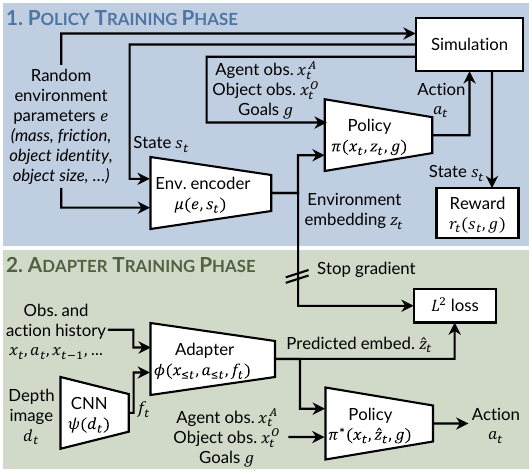}
    \caption{Overview of the proposed training procedure, which consists of 2 phases. In the first phase, a \emph{conditional policy} $\pi$ is trained to maximize a reward (e.g. move an object to a given position or orientation), given observations $x_t$ (e.g. joint angles), a goal description $g$ and privileged information about the environment $e$, $s_t$. The environment is randomized (e.g. varying mass or object identities), so an environment encoder $\mu$ is trained jointly to distill this privileged information into an embedding $z_t$. In the 2\textsuperscript{nd} phase, the policy $\pi$ and encoder $\mu$ are frozen, and an adapter $\phi$ and CNN $\psi$ are trained with a $L^2$ loss to predict the privileged information in $z_t$ from just a history of observations (e.g. past dynamic behaviour) and a depth image $d_t$ (e.g. object appearance). The adapter, CNN and policy can be deployed to perform adaptive manipulation directly from observations and depth images.}
    \label{fig:architecture}
\end{figure}

The object manipulation task is formulated as a Markov Decision Process (MDP) \cite{bellman1957markovian}. A simulator, parameterized by environment parameters $e\sim\mathcal{E}$ (e.g. robot dimensions and masses), expresses a transition probability $P_{e}$ that advances the simulation's state $s_{t}$ to the next time step $s_{t+1}$. The reinforcement learning objective is then to maximize the expected future discounted reward $r$ (measuring how well the manipulation goal is attained), when sampling trajectories $(s_{0},a_{1},s_{1},a_{2},\ldots)$ by recursive application of $P_{e}$, and taking actions $a_{t}$ that are chosen by the learned policy $\pi$:
\begin{alignat}{1}
\pi^{*}, \mu^{*}& =\underset{\pi,\mu}{\textrm{argmax}}\underset{g\sim\mathcal{G}}{\underset{e\sim\mathcal{E}}{\E}}\left[\underset{s_{t+1}\sim P_{e}(\cdot|s_{t},a_{t})}{\E}\left[\sum_{t'=0}^{T-1}\gamma^{t'}r(s_{t'},g)\right]\right]\nonumber \\
\begin{array}{c}
\textrm{with }\\
\\
\end{array} & \begin{array}{ll}
s_{0}\sim P_{e}(s_{0}), & x_{t}=o(s_{t}),\\
a_{t}=\pi(x_{t},z_t,g), & z_t=\mu(e,s_t),
\end{array}\label{eq:policy-phase}
\end{alignat}
where $T$ is the length of the simulation, $\mathcal{G}$ is a distribution over goals (e.g. a desired object position or orientation), $\mathcal{E}$ is a distribution over physical parameters (domain randomization), $0<\gamma<1$ is a discount factor to stabilize training, and $o$ is an observation function modeling the fact that the policy does not have full access to the hidden state $s_{t}$. The policy is not conditioned on the physical parameters $e$ directly, but rather on an \emph{environment embedding} $z$, which is a (possibly compressed) view of those parameters, output by an environment encoder $\mu$. It may also include other privileged (generally unobserved) information from the simulation state $s_t$. An illustration is in \cref{fig:architecture} (top half).

\cref{eq:policy-phase} is optimized using Proximal Policy Optimization (PPO) \cite{schulman2017proximal}. Both the policy $\pi$ and the environment encoder $\mu$ are multi-layer perceptrons (MLPs). Note that at this stage we have obtained a policy $\pi^{*}$ that can cope with different environment conditions, but the requirement for the environment parameters $e$ and privileged state information $s_t$ prevents its direct application in practice.

\subsection{Privileged and Observable Information}\label{sec:method-inputs}

Before moving on to the next training phase, it is worth discussing the observations $x_t$ and privileged information ($e$, $s_t$) that the policy is conditioned on.
\begin{enumerate}
    \item The \emph{observations} $x_t$ are the angle of each degree-of-freedom of the robot arm, and the position (but not orientation) of the object. Both represent measurements during deployment -- the output of proprioception (e.g. wheel encoders) and a standard object detector (e.g. vision-based).
    \item The \emph{environment parameters} $e$ represent generally unknown or hard-to-estimate quantities that are used to initialize the simulation, and are constant throughout: the manipulated object's shape, scale, mass and friction coefficient.
    \item The \emph{privileged state information} $s_t$ contains all physical variables, some of which could be useful for learning, and so we would like to encourage the model to estimate them. In our setting, we condition $\mu$ on the object's rotation in 3D, and whether there is contact on each finger (binary variables). Both are only available in simulation.
\end{enumerate}
However, these are still not enough for successful grasping, which also depends on the exact geometry of the object (\cref{sec:exp_design}).
We will address this in the following section.

\subsubsection{Category and Instance Dictionaries}\label{sec:cat-inst-dictionaries}
We propose to encode geometry only implicitly, with instance and category dictionaries of learnable embeddings as proxies for geometry knowledge. We train a dictionary of learnable embeddings (similar to word embedding vocabularies in language models \cite{bengio2003neural}), with a vector $u_i$ for the $i$th object instance, and a vector $c_j$ for the $j$th object category. These are initialized randomly, and concatenated with the physical environmental parameters $e_{\textrm{phys}}$ (\cref{sec:method-inputs}):
\begin{equation}
    e(i)=\left(e_{\textrm{phys}}(i),u_{i},c_{\textrm{cat}(i)}\right)\label{eq:env-embed}
\end{equation}
where cat$(i)$ retrieves the index of the category of object instance $i$.
While this information is not available during deployment, it is no different from the other privileged physical parameters $e_{\textrm{phys}}$. This encoding allows us to estimate both kinds of privileged information with the same method, presented in the next section.

\subsection{Adapter Training Phase}\label{sec:adapter}

In order to estimate the environment embedding $z$ with readily-available information, instead of privileged environment parameters $e$, the second phase aims to train an \emph{adapter }$\phi$ that is conditioned on past observations $x_{\leq t}$ and actions $a_{\leq t}$. Note that it is unlikely that observations based purely on proprioception (joint angles) will carry information about objects' identities prior to manipulating them (\cref{eq:env-embed}). This necessitates another input modality, to allow conditioning on object categories and instances (albeit indirectly). We choose depth images $d_{t}$ from an arm-mounted camera, which should reveal properties such as an object's size or the orientation of graspable features, such as handles. These are processed by a convolutional neural network (CNN) $\psi$ before being passed to the adapter $\phi$. The overall objective then becomes:
{
\thinmuskip=1mu
\medmuskip=2mu
\thickmuskip=3mu
\begin{alignat}{1}
\phi^{*},\psi^{*} & =\underset{\pi,\psi}{\textrm{argmin}}\underset{g\sim\mathcal{G}}{\underset{e\sim\mathcal{E}}{\E}}\left[\underset{\substack{s_{t+1}\sim\quad\;\;\;\;\\
\;\, P_{e}(\cdot|s_{t},a_{t})
}}{\E}\left[\sum_{t'=0}^{T-1}\left\Vert \mu^{*}(e,s_{t'})-\hat{z}_{t'}\right\Vert ^{2}\right]\right]\nonumber \\
\begin{array}{c}
\textrm{with }\\
\\
\end{array} & \begin{array}{lll}
s_{0}\sim P_{e}(s_{0}), & a_{t}=\pi^{*}(x_{t},\hat{z}_t,g), & x_{t}=o(s_{t}),\\
f_{t}=\psi(d_{t}), & \hat{z}_t=\phi(x_{\leq t},a_{\leq t},f_{t}),
\end{array}\label{eq:adapter-phase}
\end{alignat}
}
where $\hat{z}_t$ is the estimated environment embedding. Note that the optimal policy $\pi^{*}$ and environment encoder $\mu^{*}$ from the first phase are used but kept frozen (i.e. not minimized over).
An illustration is in \cref{fig:architecture} (bottom half).

\cref{eq:adapter-phase} is optimized using standard back-propagation (namely Adam \cite{kingma2014adam}). During deployment, only the trained depth CNN $\psi^{*}$, adapter $\phi^{*}$ and policy $\pi^{*}$ are used.

\subsection{Environments} 
We train our agents in a customized variant of \emph{ManiSkill2} environments \citep{gu2023maniskill2}, with additional environmental randomization (\cref{sec:method-inputs}).
We show an illustration of each task in \cref{fig:trajectories}.
These tasks are:
\begin{enumerate}
    \item Pick and Place YCB and EGAD objects. The agent picks up a random object from the YCB dataset~\citep{calli2015ycb} (78 objects), or the EGAD dataset~\cite{morrison2020egad} (2281 objects), and places it at a point uniformly sampled from the reachable 3D space.
    \item Peg Insertion. The agent picks up a cuboid-shaped peg on the table, and inserts at least 50\% of it into a gap.
    \item Faucet Turning. The agent turns a faucet handle by a variable angle, with a random faucet from the 60-object PartNet-Mobility dataset \cite{xiang2020sapien}.
\end{enumerate}

The selected tasks exemplify a broad spectrum of goal specifications.
The first two have only a positional target, while Peg Insertion has both positional and rotational specifications.
It also has partial constraints on the moving trajectory for the peg to be successfully inserted into the hole, rather than simply matching a target pose.
Faucet Turning requires rotating (to varied angles) a faucet handle (of varied shapes; see \cref{app:exampl_objs} for examples).
This skill is representative of other useful ``twisting'' motions, such as rotating screwdrivers, or unscrewing caps to open containers.

We use the default task-specific dense rewards offered by \emph{Maniskill2} environments \citep{gu2023maniskill2}, which are composed of simple metrics such as distances between entities or whether the object is grasped for training the agents (see \cref{app:reward_functions} for an overview).

\section{Experiment Design}\label{sec:exp_design}

\paragraph{Simulation Setup.}
We use the Franka Emika Panda robot arm, a widely used 7-DOF manipulator with torque sensors in each joint known for its dexterity and precision. 
The arm is controlled using position control at a frequency of 20 Hz. 
Complementing the arm is a two-finger gripper, which serves as the end-effector for object manipulation tasks.
To convert target position commands into actuator torques, we utilize a Proportional-Derivative (PD) controller with stiffness and damping coefficients \( K_p = 4.0 \) and \( K_d = 0.2 \), respectively.
These can also be varied and added to the list of environment variation parameters in future work.
 
We utilize the ManiSkill2 environments \citep{gu2023maniskill2} constructed atop the Sapien simulator \citep{xiang2020sapien}.
During training, 50 independent environments run concurrently, where each episode has a length of 50 control steps; in testing, each episode has a maximum length of 200 control steps.
The simulation operates at a frequency of 120 Hz while the control policy operates at 20 Hz.
This setup provides a robust and versatile platform for evaluating the performance of our algorithms, and for extending it to mobile robots and to real-world manipulation tasks in future work.

\paragraph{Environment Setup.}
As introduced on a high level in \cref{sec:method}, we incorporate three types of randomization into each environment for learning a generalizable policy.
In each run, the parameters are sampled from a uniform distribution parameterized by the boundary values (see \cref{app:env_params}).
For evaluation of agent generalization, the ranges of environmental variations, observation noise, and external disturbances are widened during testing.
Specifically, we increased the low and high values of the environment variations and external disturbance distribution by 0.8 and 1.2 during testing, respectively.
We scale both the boundary parameters for the observation noise distribution by 1.2.

External disturbances are forces applied onto an object's center of mass when it is grasped by the robot.
Following \citep{andrychowicz2020learning, qi2023hand}, we implement it as follows.
At each control step, we sample from a Bernoulli distribution with probability $p$ whether to apply such a force to the object. If true, we apply a randomly sampled force to the object, which is then decayed by 0.8 at each control step.
To sample the force to be applied, we first sample a direction vector from a 3-dimensional Gaussian distribution with mean 0, standard deviation 0.1, and scaled it to have an L2 norm of 1. 
The force is then scaled by the object mass and a force scale parameter sampled (see \cref{app:env_params}).
When a new force is sampled, the residual force from the previous time step is overwritten.

Differing tasks naturally yield variations in the privileged information $e_t$ and the goal state $g_t$, while the object and agent state shapes are the same across tasks.
Specifically, all three tasks share the agent state $x_t^a \in \mathbb{R}^{32}$ which includes the 9-dimensional position and velocity of its joints, and the 7-dimensional pose of its base and Tool Center Point (TCP).

The object state $x^o_t \in \mathbb{R}^{6}$ is a concatenation of the object position, and $\|^\text{tcp}x_t^\text{obj}\|$ -- the distance between the TCP and the object center.
We assume the object position is output by an off-the-shelf perception module with imperfect accuracy.
Alternatively, this could also be part of the privileged information whose embeddings can be estimated from the depth-based perception in phase 2 of the training.

The privileged environment information $e_t\in\mathbb{R}^{71}$ is a concatenation of 
object dimension, $e_t^{\text{dim}}\in\mathbb{R}^{3}$; 
object density, $e_t^{\text{dens}}\in\mathbb{R}$; 
friction coefficient, $e_t^{\text{fric}}\in\mathbb{R}$; 
the magnitude of the impulse applied by the left and right finger of the gripper $e_t^{\text{impl}}\in\mathbb{R}^2$; 
and a 64-dimensional embedding for the type and token variable for the object identity, $e_t^{\text{typ}}, e_t^{\text{tok}}\in\mathbb{R}^{32}$.
For Faucet Turning, the object dimension is replaced with the rotational axis of the handle that is targeted by the task to rotate $e_t^{\text{axis}}\in\mathbb{R}^{3}$.

The goal state representation naturally varies by task.
In Pick and Place with YCB and EGAD objects, $g_t\in\mathbb{R}^{9}$ consists of the target 3-dimensional position of the object $g_t^{\text{pos}}\in\mathbb{R}^{3}$ and $^\text{tcp}x^\text{goal}_t, ^\text{obj}x^\text{goal}_t \in \mathbb{R}^3$ which refers to the distance in position between the TCP and the goal, and the object and the goal, respectively.
The derived variables assist the policy by extracting useful information that guides actions, which simplifies the task of learning.
For Peg Insertion, the goal $g_t\in\mathbb{R}^{13}$ includes the pose of the target hole $g_t^{\text{pos}}\in\mathbb{R}^{7}$. 
And it also contains the $^\text{tcp}x^\text{goal}_t, ^\text{obj}x^\text{goal}_t \in \mathbb{R}^3$ as in Pick and Place.
For Faucet Turning, the goal $g_t\in\mathbb{R}^2$ specifies the 1-dimensional angle to rotate the handle with. 

We train separate policies for each of the tasks to explore the feasibility and generalization of single-task agents.
In future works, we wish to explore the possibility of multi-task agents with the hope that knowledge about dexterous movements can be shared across different tasks, accelerating the learning process.

\paragraph{Baselines and Ablations.}
We compare our model, dubbed \textit{RMA$^2$}, against the following ablations and baselines.
Each comparison is designed to highlight a different aspect of our design.
The alternative models include:

\begin{enumerate}
\item \textbf{Oracle Adaptation} (\textit{Oracle}). This model uses the ground truth extrinsic vector $z_t$ generated by the environment encoder as opposed to the estimated $\hat{z}_t$. 
Because it relies on ground-truth access to privileged information, this alternative model could never actually run in the real world, but serves as an upper bound on adaptation performance.

\item \textbf{Domain Randomization with state-based policy} (\textit{DR}).
This baseline implements basic domain randomization, trained using the same randomization scheme but without the privileged information \cite{tobin2017domain}. 
This comparison serves to test the value of adaptation, by replacing it with a policy that does not adapt but aims to be robust across environment variations.

\item \textbf{Domain Randomization with vision-based policy} (\textit{DR+Vi}).
This baseline uses depth-based perception rather than state-based info (as for \textit{DR}), similar to \citep{andrychowicz2020learning}\footnote{We also experimented with DR+Vi+Proprioception but achieved similar performance as DR+Vi.}.
\item \textbf{Automatic Domain Randomization} (\textit{ADR}).
This baseline uses ADR to generate learning curricula for improved efficiency, as done in a number of recent works \citep{handa2023dextreme, akkaya2019solving}.
\item \textbf{Without Object Embedding} (\textit{NoOE}).
This model omits the two-part object embedding during training and, as a result, remains unaware of the identity of the object being manipulated.
This variation assesses the benefit of incorporating object type-token identity into the privileged information.

\item \textbf{No Vision in Adaptation} (\textit{NoVA}).
This ablation removes depth vision when predicting extrinsics $\hat{z}_t$, similar to the adaptation in previous RMA works \citep{kumar2021rma, qi2023hand}.

\end{enumerate}

\paragraph{Metrics.}
We adopt the following metrics to evaluate the models, each giving a different lens on model performance.
The results, based on these metrics, are averaged across three random seeds, with each seed's result averaged across 5000 episodes:

\begin{enumerate}
\item \textbf{Success Rate} (SR). This gauges the proficiency of the agent in performing the assigned task. 
An episode is deemed successful if the agent meets the task's objective as defined in ManiSkill2~\cite{gu2023maniskill2}. For example, for the Pick and Place task, success is attained when the object is positioned within 2.5 cm of the target location, with the robot remaining static.

\item \textbf{Episode Length} (EL). This measures the time taken by the agent to complete the task, with a cap set at 200 steps.
Shorter length signifies a more efficient task completion.
\end{enumerate}

\paragraph{Architecture and Training Details.}
In policy training, we use a 3-layer MLP for Environment encoder and a 4-layer MLP for Policy.
In adapter training, we use a CNN with 3 2D convolutional layers and 4 1D convolutional layers for the depth image and state-action history, respectively.
This sensory information is then integrated by a 2-layer MLP in Adapter.
The parameters are optimized with the Adam optimizer \cite{kingma2014adam}.

We use a curriculum learning approach to facilitate the learning process. This curriculum linearly amplifies the magnitude of three types of randomization in our environment--environment variations, 
external disturbances, and observation noise, up to a threshold.

We train each agent on an Nvidia A100 GPU and 16 CPUs until convergence, or a maximum of 7 days.

\begin{table*}[ht]
\small

\begin{tabular*}{1\textwidth}{@{\extracolsep{\fill}}ccccccccc}
\toprule 
 & \multicolumn{2}{c}{Pick \& Place task (YCB)} & \multicolumn{2}{c}{Faucet Turning task} & \multicolumn{2}{c}{Peg Insertion task} & \multicolumn{2}{c}{Pick \& Place task (EGAD)}\\
\midrule 
Method & SR $\uparrow$ & EL $\downarrow$ & SR $\uparrow$ & EL $\downarrow$ & SR $\uparrow$ & EL $\downarrow$ & SR $\uparrow$ & EL $\downarrow$\\
\midrule
\textcolor{gray}{\emph{Oracle}} & \textcolor{gray}{$75.4\pm0.6$} & \textcolor{gray}{$64.2\pm1.4$} & \textcolor{gray}{$76.2\pm0.4$} & \textcolor{gray}{$70.0\pm0.3$} & \textcolor{gray}{$55.4\pm5.6$} & \textcolor{gray}{$111.9\pm10.4$} & \textcolor{gray}{--} & \textcolor{gray}{--}\\
\midrule
DR & $0.3\pm0.1$ & $199.7\pm0.0$ & $48.9\pm3.1$ & $176.1\pm4.5$ & $45.6\pm21.6$ & $142.2\pm34.3$ & $1.0\pm0.0$ & $199.9\pm0.0$\\
DR+Vi & $35.2\pm3.7$ & $138.1\pm6.6$ & $14.7\pm3.2$ & $169.6\pm3.8$ & $0.0\pm0.0$ & $200.0\pm0.0$ & $35.5\pm4.5$ & $138.6\pm7.2$\\
ADR & $1.6\pm2.3$ & $198.1\pm2.1$ & $51.8\pm1.8$ & $110.1\pm2.5$ & $14.0\pm19.0$ & $177.6\pm30.5$ & $3.0\pm3.5$ & $196.4\pm4.3$\\
NoOE & $70.4\pm0.4$ & $77.7\pm2.9$ & $57.6\pm0.4$ & $108.3\pm1.0$ & -- & -- & $88.1\pm0.1$ & $44.3\pm0.9$\\
NoVA & $68.1\pm0.8$ & $84.4\pm0.5$ & $47.0\pm0.7$ & $182.6\pm1.8$ & $48.4\pm9.8$ & $133.6\pm43.9$ & $87.5\pm3.4$ & $52.6\pm8.8$\\
\midrule 
\textbf{RMA$^{2}$} & $\mathbf{73.8\pm4.5}$ & $\mathbf{72.1\pm0.8}$ & $\mathbf{62.7\pm0.5}$ & $\mathbf{88.1\pm0.8}$ & $\mathbf{51.6\pm6.7}$ & $\mathbf{127.3\pm10.8}$ & $\mathbf{90.5\pm2.8}$ & $\mathbf{40.1\pm6.1}$\\
\bottomrule
\end{tabular*}
\caption{Evaluation results of our model and the baselines in simulation for the 4 tasks that we evaluate.
The \textit{Oracle} agent is highlighted in gray, as it is not real-world applicable due to its reliance on privileged information.
The methods are Domain Randomization (DR)~\cite{tobin2017domain}, a reactive vision-based RL method (DR+Vi)~\cite{andrychowicz2020learning}, Automatic Domain Randomization (ADR)~\cite{handa2023dextreme, akkaya2019solving}, an ablation of our method without object embeddings (NoOE), and our method without depth vision in the adapter (NoVA), i.e. simple RMA~\cite{kumar2021rma, qi2023hand}.
The best performance in each column is bolded.
See \cref{sec:result} for more details.}
\label{tab:exp_result}
\end{table*}

\section{Experiment Results and Analysis}\label{sec:result}
We show example trajectory of \textit{RMA}$^2$ in each task in \cref{fig:trajectories} and the experiment results of it and the baselines in \cref{tab:exp_result}.
We see that \textit{Oracle} consistently achieves the highest success rate and lowest episode length across the tasks, which is expected given its privileged access to the simulation's parameters and state.
Our method, \textit{RMA}$^2$, is consistently the best performing agent in the evaluation while being real-world deployable.
The two ablations, \textit{NoOE} and \textit{NoVa} closely follow \textit{RMA}$^2$'s performance.
This highlights the significance of each of the design choices.
Their performance is sometimes better than the agents trained with domain randomization, but not always -- both object dictionaries and depth conditioning are necessary to achieve the best result.

\subsection{Pick \& Place task -- YCB objects dataset}

At the task of picking and placing objects sampled from the YCB dataset (see \cref{fig:trajectories}, row 1 for an example), \textit{DR} and \textit{ADR} exhibit a negligible success rate in the allotted time, underscoring our method's proficiency in reducing sample complexity and enhancing practical task learnability (see \cref{tab:exp_result}, columns 2 and 3).
The vision-based \textit{DR+Vi} achieves better results than \textit{DR} but is about 4 times slower than the state-based method to complete the same number of training steps, such as \textit{RMA}$^2$ and \textit{DR}.
The ablation \textit{NoVA} outperforms \textit{NoOE}, showing the value of vision-based adaptation.

\subsection{Faucet Turning task}

In the Faucet Turning task (see \cref{fig:trajectories} row 2 for an example trajectory), \textit{ADR} outperforms \textit{DR}, benefiting from the dynamically generated curriculum (\cref{tab:exp_result}, column 4 and 5).
\textit{DR+Vi} did not converge and scored lower than \textit{DR} after 7 days of training, as it was much more computationally-expensive to train.
There is a larger gap between \textit{RMA}$^2$ and \textit{Oracle} than in Pick and Place with YCB objects ($1.6\%$ vs $13.5\%$), which is reflected in the larger adaptation loss ($.08$ vs $.04$).
We hypothesize that the movement of the gripper camera increases training complexity, which could potentially be addressed with a fixed camera, but this might bring additional challenges in ``hand-eye'' coordination.

\begin{figure*}[htbp]
  \centering
  \begin{subfigure}{.3\linewidth}
    \centering
    \includegraphics[width=.99\linewidth]{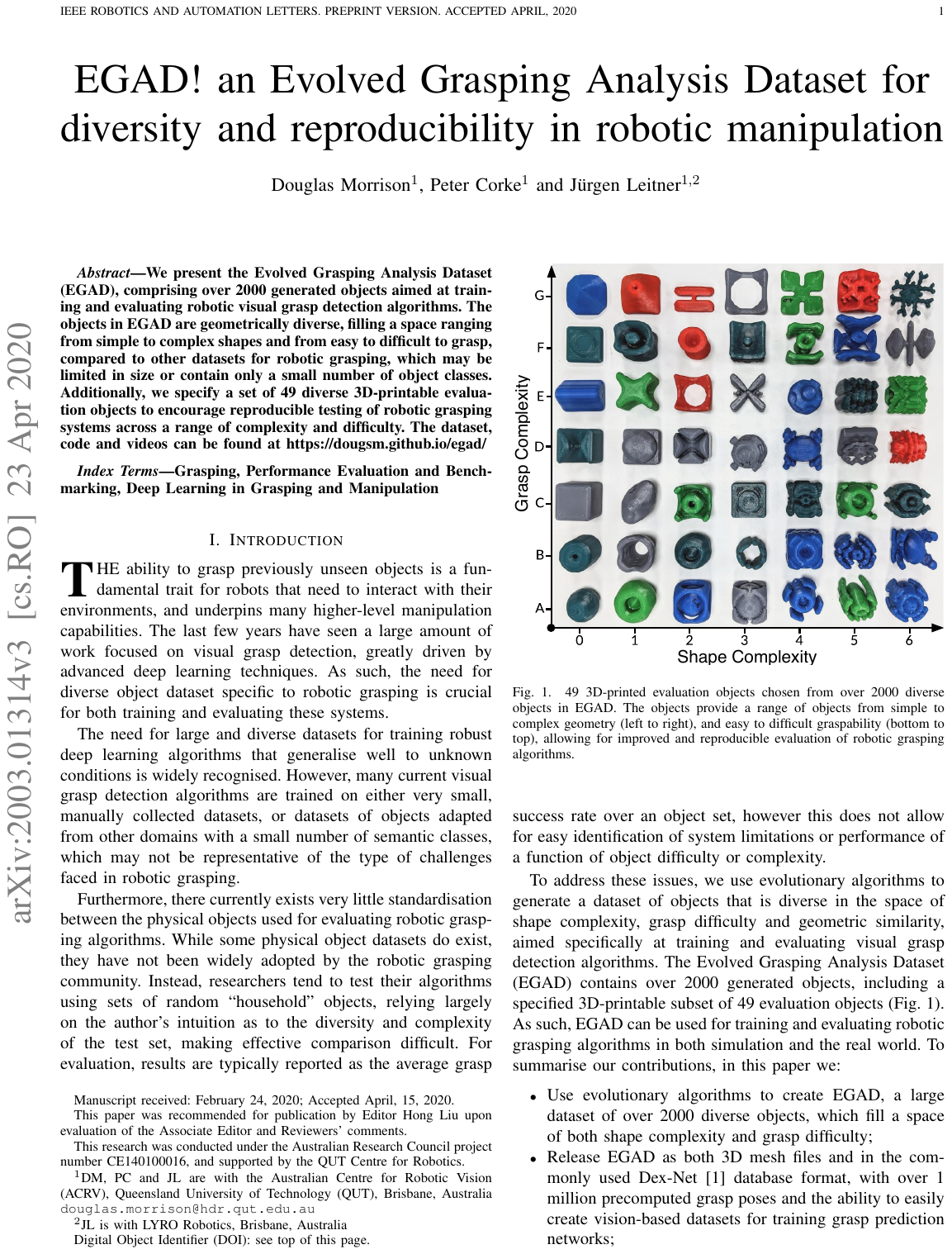}
    \caption{}
  \end{subfigure}%
  \begin{subfigure}{.72\linewidth}
    \centering
    \includegraphics[width=.99\linewidth]{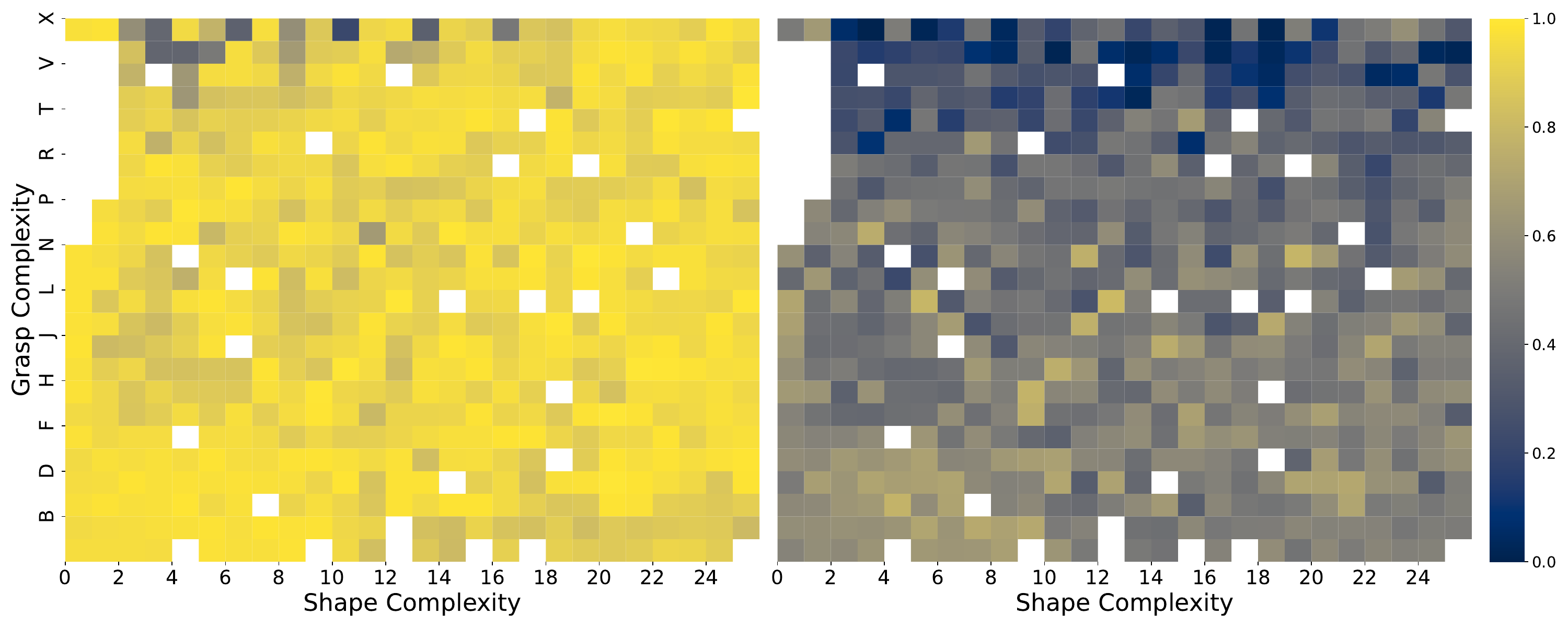}
    \caption{}
    \label{fig:sub4}
  \end{subfigure}
  \caption{
  (a) Example objects from the EGAD dataset, sorted by grasp and shape complexity. This illustrates the array of diverse shapes.
  The horizontal axis indicates ascending shape complexity, while the vertical axis corresponds to increasing grasp complexity. 
  (b) Fine-grained evaluation of the performance of \textit{RMA}$^2$ (left) and \textit{DR+Vi} (right) on Maniskills2's Pick \& Place task, with EGAD objects.
  The color coding reflects the success rate (bright yellow for 100\%, dark blue for 0\%), averaged over 500 runs.
  The white cells corresponds to objects that are not in the dataset for this task.
  }
  \label{fig:egad_heatmap}
\end{figure*}

\subsection{Peg Insertion task}

Overall, Peg Insertion is the most challenging task as the hole has only a 3 mm clearance on the box and the task is successful only if half of the peg is inserted, while the equivalent task in other benchmarks~\cite{yu2020meta} only requires the peg head to approach the surface of the hole (refer to the third row of \cref{fig:trajectories} for a sample trajectory). 
In this task, we removed the \textit{NoOE} ablation as there is only one object shape, the cuboid-shaped peg, in the task.

\textit{ADR} performs worse than \textit{DR}, which likely indicates a suboptimal hyperparameter setting for this task (\cref{tab:exp_result}, column 6 and 7).
\textit{DR+Vi} achieves $0.0\%$ accuracy again due to it reaching the timeout before making any progress in the task, indicating that direct visual policy learning may be too difficult when the task requires high precision to even receive a reward.

\subsection{Extrapolation of Policies from YCB to EGAD Dataset}

EGAD is a collection of more than 2000 geometrically unique object generated using evolutionary algorithms specifically for evaluating robotic grasping and manipulation~\cite{morrison2020egad} (see \cref{fig:egad_heatmap}~(a) for an illustration and \cref{app:exampl_objs}).
We evaluated the agents trained on the YCB dataset directly on this to evaluate the effect of shift in object shape distribution, with an example trajectory shown in \cref{fig:trajectories} row 4.

The \textit{Oracle} agent is not applicable here because the trained object type-token embeddings are available only for the objects it has been trained on.
Notably, as tabulated in the last two columns of \cref{tab:exp_result}, \textit{RMA}$^2$'s success rate is higher than its performance on the YCB dataset by a $16.7\%$ margin, while \textit{DR+Vi}'s performance is only $0.3\%$ higher than its counterpart on the YCB objects.
This highlights the greater generalization performance for our method compared to domain randomization methods, which require the randomization distribution to be well-tuned to the distribution of the task that the policy is ultimately deployed to.

In \cref{fig:egad_heatmap}~(b), we present the per object success rate for \textit{RMA}$^2$ and \textit{DR+Vi}.
Overall, the heatmap for \textit{RMA}$^2$ exhibits a consistently brighter tone, indicating a generally higher success rate.
Notably, the \textit{RMA}$^2$ heatmap shows a pronounced darkness in areas where shape complexity is low yet grasp complexity is high, which is not as apparent when both complexities are elevated.
In contrast, \textit{DR+Vi} demonstrates a darker region across the spectrum of high grasp complexity, indicating a more uniform challenge in these conditions.
Despite both approaches utilizing a CNN, we hypothesize that the deliberate inductive bias in \textit{RMA}$^2$, which only attempts to predict an environment embedding that is useful for conditioning a successful policy, allows it to generalize better to geometrically more complex shapes.

\section{Conclusion}\label{sec:concludsion}
In this work, we presented Rapid Motor Adaptation for Robot Manipulator Arms (\textit{RMA}$^2$).
By incorporating a category-instance dictionary, paying deliberate attention to environmental parameters in base policy training and utilizing low-resolution depth vision during adaptation training, our policy demonstrated superior generalization performance and sample efficiency across four challenging ManiSkill2 tasks compared to the baselines.
We believe these principles can be leveraged for efficient learning of other complex manipulation skills.

Looking ahead, we see several promising avenues for further research.
\begin{inparaenum}[1)]
\item A natural next step is to learn a multitask motor skill policy that encourages knowledge sharing across an even broader range of tasks, which could further improve adaptability and learning efficiency.
\item Building on the state-based observations,
an interesting extension would be to support variable numbers of objects in the environment.
\item We observe in tasks such as faucet turning that there is a performance gap between \textit{Oracle} and \textit{RMA}$^2$,
which suggests that there is room for improving the adapter's estimate of the environment embedding, potentially by including other modalities or more sophisticated visual networks.
\item Finally, low-level skills could seamlessly interoperate with high-level task planners, or hierarchical RL methods, to develop more versatile and adept embodied AI agents capable of achieving long-horizon tasks in the real world.
\end{inparaenum}

\paragraph{Acknowledgements}
The authors would like to express their gratitude to Doug Morrison for the permission to reuse their figure. 
We acknowledge the generous support of the Royal Academy of Engineering (RF\textbackslash 201819\textbackslash 18\textbackslash 163).
YL is particularly thankful for the funding provided by the Cambridge Trust and an unrestricted gift from Huawei.

{
    \small
    \bibliographystyle{ieeenat_fullname}
    \bibliography{main}
}

\newpage

\setcounter{page}{1}
\maketitlesupplementary
\appendix
\section{Environment Randomization Details}\label{app:env_params}

The sample range for the environment parameters during training and evaluation is shown in \cref{tab:env_params}.

\begin{table}[h]
\centering
\small %
\begin{tabular}{llll}
\hline
\textbf{Rand. type} & \textbf{Params.} & \textbf{Training} & \textbf{Testing} \\
\hline
\multirow{4}{*}{Env. Var.} & Scale & [0.70, 1.20] & [0.56, 1.44]\\
\cline{2-4}
 & Density & [0.50, 5.00] & [0.40, 6.00]\\
\cline{2-4}
 & Friction & [0.50, 1.10] & [0.40, 1.32]\\
\cline{2-4}
 & Shape & All & All$^*$ \\
\hline
\multirow{1}{*}{Ex. Dist.} & Force & [0.00, 2.00] & [0.00, 2.40] \\
\hline
\multirow{3}{*}{Obs. Nois.} & Obj. Pos. & [-0.005, 0.005] & [-0.006, 0.006]\\
\cline{2-4}
 & Obj. Rot. & [-10, 10] & [-12, 12]\\
\cline{2-4}
 & Joint Pos. & [-0.005, 0.005] & [-0.006, 0.006]\\
\hline
\end{tabular}
\caption{
Summary of environment randomization for training and testing phases. The table specifies the range of uniform distribution from which the parameters are sampled. 
Environment Variations (Env. Var.) encompass object properties such as object scale multiplier (which is multiplied on top of the default object scale), object density multiplier, object coefficient of friction, and object shape.
$^*$The pick and place agent is trained on all the YCB objects and evaluated on both YCB and EGAD objects.
The faucet turning agent is trained and tested on the same set of faucets from the PartNet-Mobility dataset.
External Disturbance (Ex. Dist.) simulates environmental perturbations through force applied onto the object. 
Observation Noise (Obs. Nois.) accounts for inaccuracies in the object position, rotation estimation and joint position reading.}
\label{tab:env_params}
\end{table}

\section{Example Objects from the Tasks}\label{app:exampl_objs}

\begin{figure}[htbp]
    \centering
    \begin{subfigure}[b]{0.49\textwidth}
        \includegraphics[width=\textwidth]{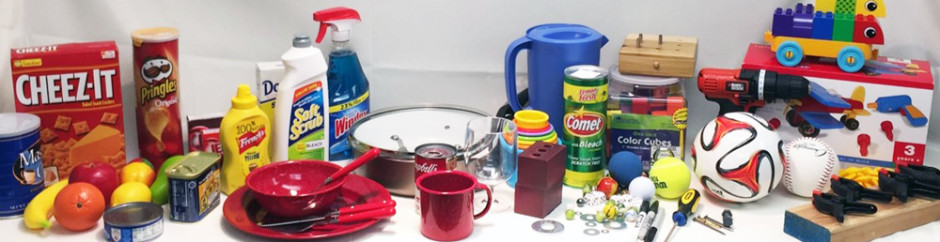}
        \caption{Objects in the YCB dataset}
    \end{subfigure}
    \vspace{-.095em}
    \begin{subfigure}[b]{0.49\textwidth}
        \includegraphics[width=\textwidth]{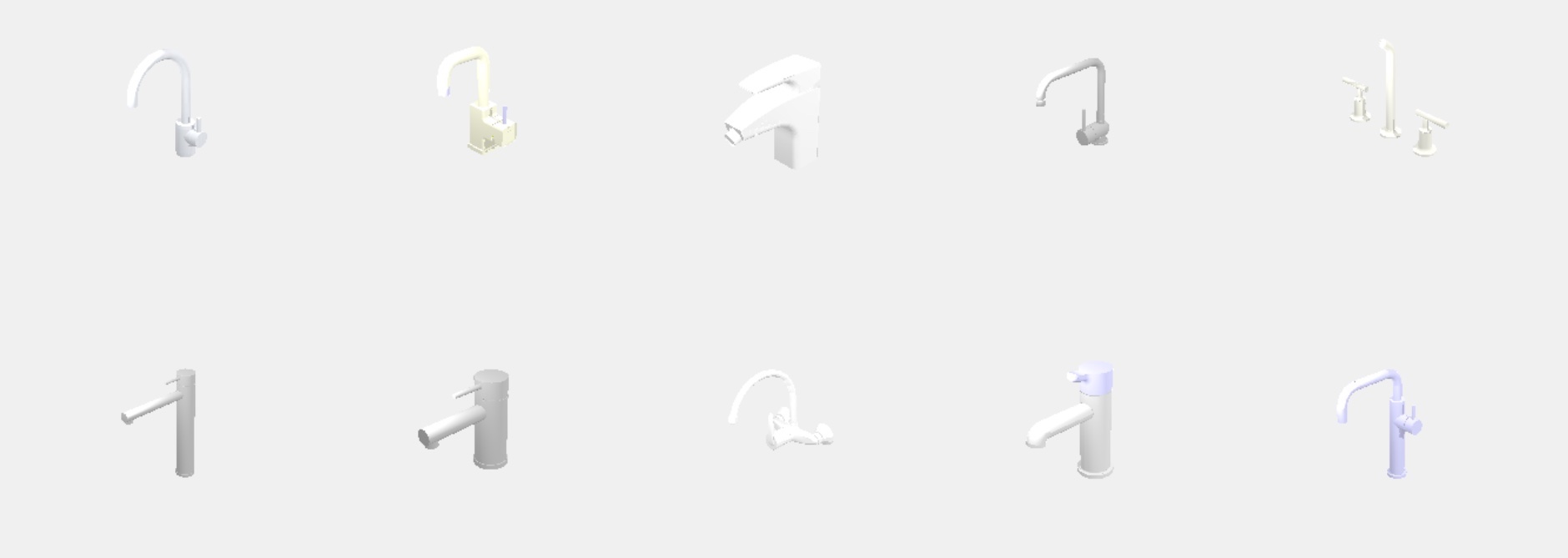}
    \end{subfigure}
    \begin{subfigure}[b]{0.49\textwidth}
        \includegraphics[width=\textwidth]{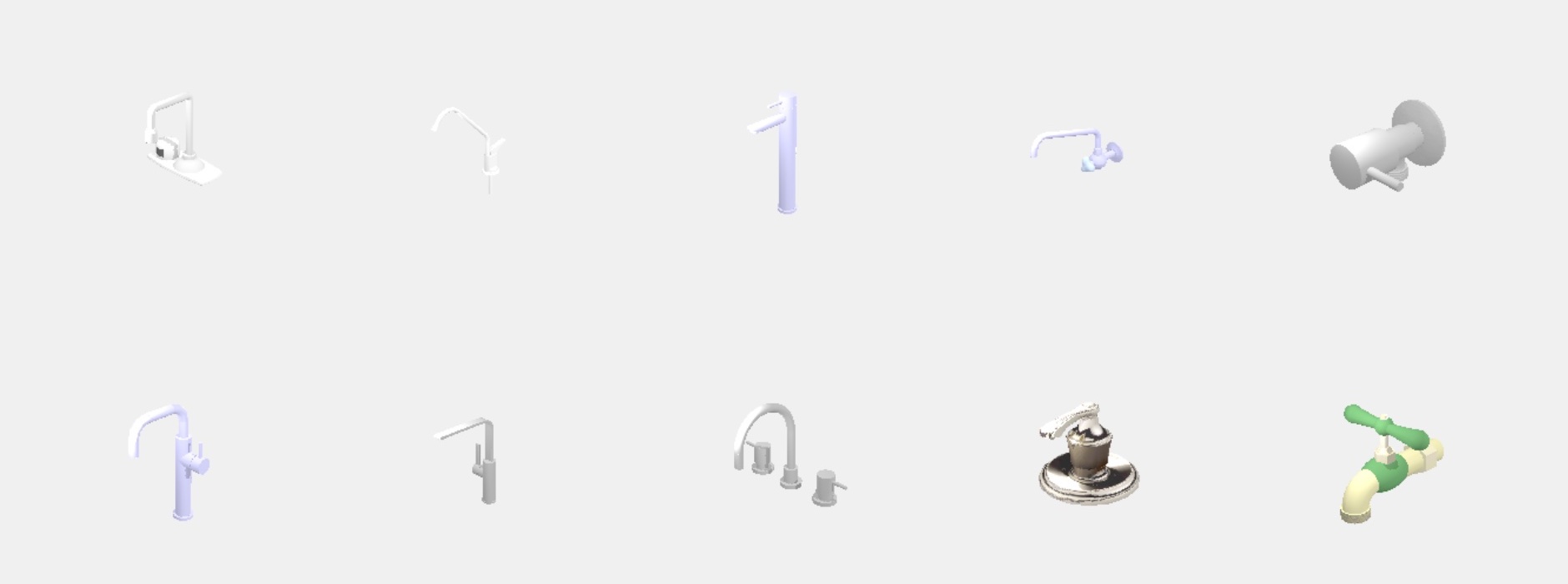}
        \caption{Sample Faucets}
    \end{subfigure}
    \caption{
    Figure (a) illustrates the YCB object we use in the Pick and Place task.
    Figure (b), (c) shows sample faucets for the Faucet Turning task.}
    \label{fig:objs}
\end{figure}

\cref{fig:objs} shows example objects from the YCB object dataset\footnote{Figure reproduced from \url{www.ycbbenchmarks.com} with permission.}\cite{calli2015ycb} and Faucets from the PartNet-Mobility dataset used in the Faucet Turning task\cite{xiang2020sapien}.
For EGAD objects, shape complexity is measured using the entropy of a histogram based on the angular deficit at each vertex of a 3D mesh. 
Grasp difficulty is estimated using the 75th percentile of the Ferrari-Canny quality metric calculated from sampled antipodal grasps on each object 

\section{Rewards}\label{app:reward_functions}
We use the default dense reward functions from Maniskill2~\cite{gu2023maniskill2}.
We give an overview of the reward function used in Pick and Place.
And we refer to the reward functions for Faucet Turning and Peg Insertion to ManiSkill2's repository\footnote{\url{github.com/haosulab/ManiSkill2}}.

For Pick and Place, the objective is to move the objects to be within 2.5 cm of the goal position, and the robot is static.
Once the goal is achieved, the agent receives a reward of 10. Otherwise, the reward at time step is the sum of the following three components:
\begin{equation}
r_t =
\begin{cases} 
10 & \text{if succ}_t, \\
1 - \tanh(3 \Delta_t) + 3 \cdot \mathbbm{1}_{\text{is\_grasped, t}} \\ 
\quad + 3 \cdot (1 - \tanh(3 \|^{\text{obj}}\mathbf{x}^\text{goal}_t\|)) & \text{o/w}.
\end{cases}
\end{equation}
where \( \text{succ}_t \) is a binary indicator of success at time step \( t \), which equals 1 if the task is successfully completed, otherwise 0;
$\Delta_t = \max(\|^{\text{tcp}}\mathbf{x}_t^{\text{obj}}\| - \|\mathbf{d}_{\text{bbox}}\|, 0)$ represents the distance between the tool center point (TCP) of the robot's end effector and the object at time step \( t \), after accounting for the bounding box dimensions of the object; 
\(\|^{\text{tcp}}\mathbf{x}_t^{\text{obj}}\| \) is the Euclidean norm representing the distance between the TCP and the object at time step \( t \);
\( \|\mathbf{d}_{\text{bbox}}\| \) is the Euclidean norm representing the dimensions of the object's bounding box;
\( \mathbbm{1}_{\text{is\_grasped, t}} \) is an indicator function that equals 1 if the object is grasped at time step \( t \), otherwise 0;
\( \|^{\text{obj}}\mathbf{x}^\text{goal}_t\| \) is the Euclidean norm measuring the distance between the object and the goal location at time step \( t \).

\section{Sample Evaluation Videos}
We showcase an example trajectory of our method, RMA$^2$, vs. the domain randomization baseline from the evaluation of each of the four tasks and a lemon-picked representative failure episode from RMA$^2$ at \url{https://youtu.be/Vii6kx3-sTQ}.
We found the most common failure mode is a failure to insert the peg precisely into the hole under disturbances, or a failed initial grasp (in pick and place and faucet turning), likely due to uncertainty about the object geometry under RL training.
\end{document}